\documentclass[10pt,twocolumn,letterpaper]{article}

\usepackage{iccv}
\usepackage{times}
\usepackage{epsfig}
\usepackage{graphicx}
\usepackage{algorithm}
\usepackage{amsmath}
\usepackage{amssymb}
\usepackage{caption}
\usepackage{float}
\usepackage{mathtools}
\usepackage{subcaption}
\usepackage{listings}
\usepackage[dvipsnames]{xcolor}
\usepackage[noend]{algpseudocode}

\DeclarePairedDelimiter\norm{\lVert}{\rVert}%

\definecolor{codegreen}{rgb}{0,0.6,0}
\definecolor{codegray}{rgb}{0.5,0.5,0.5}
\definecolor{codepurple}{rgb}{0.58,0,0.82}
\definecolor{backcolour}{rgb}{0.95,0.95,0.92}
 
\lstdefinestyle{mystyle}{
    backgroundcolor=\color{backcolour},   
    commentstyle=\color{codegreen},
    keywordstyle=\color{magenta},
    numberstyle=\tiny\color{codegray},
    stringstyle=\color{codepurple},
    basicstyle=\ttfamily\footnotesize,
    breakatwhitespace=false,         
    breaklines=true,                 
    captionpos=b,                    
    keepspaces=true,                 
    numbers=left,                    
    numbersep=5pt,                  
    showspaces=false,                
    showstringspaces=false,
    showtabs=false,                  
    tabsize=2
}

\newfloat{lstfloat}{htbp}{lop}
\floatname{lstfloat}{Listing}

\usepackage[pagebackref=true,breaklinks=true,colorlinks,bookmarks=false]{hyperref}

\iccvfinalcopy 

\def\iccvPaperID{11} 

\ificcvfinal\fi

\begin{document}

\title{Deep Closed-Form Subspace Clustering}

\author{Junghoon Seo\\
Satrec Initiatve, Republic of Korea\\
{\tt\small sjh@satreci.com}
\and
Jamyoung Koo \hspace{0.5cm} Taegyun Jeon\\
SI Analytics, Republic of Korea\\
{\tt\small \{jmkoo,tgjeon\}@si-analytics.ai}
}

\maketitle
\ificcvfinal\thispagestyle{empty}\fi

\begin{abstract}
We propose \textbf{D}eep \textbf{C}losed-\textbf{F}orm \textbf{S}ubspace \textbf{C}lustering (\textbf{DCFSC}), a new embarrassingly simple model for subspace clustering with learning non-linear mapping.
Compared with the previous deep subspace clustering (DSC) techniques, our DCFSC does not have any parameters at all for the self-expressive layer.
Instead, DCFSC utilizes the implicit data-driven self-expressive layer derived from closed-form shallow auto-encoder.
Moreover, DCFSC also has no complicated optimization scheme, unlike the other subspace clustering methods.
With its extreme simplicity, DCFSC has significant memory-related benefits over the existing DSC method, especially on the large dataset.
Several experiments showed that our DCFSC model had enough potential to be a new reference model for subspace clustering on large-scale high-dimensional dataset.
\end{abstract}

\section{Introduction}
In this paper, we tackle the problem of subspace clustering on high-dimensional and large-scale dataset.
Subspace clustering \cite{vidal2011subspace} seeks to find clusters in the dataset by selecting the most relevant dimensions for each cluster separately.
It has become an import topic in unsupervised learning and achieved great success in various computer vision tasks, such as face clustering \cite{ho2003clustering}, image segmentation \cite{yang2008unsupervised}, and motion segmentation \cite{ji2016robust, rao2008motion}.

Recently, methods on subspace clustering based on sparse and low-rank representation \cite{elhamifar2013sparse, you2016scalable, liu2012robust, wang2013provable, patel2015latent, zhang2015low} have gotten attention.
Many of these methods exploited self-expressiveness property \cite{rao2008motion, elhamifar2009sparse} of data drawn from a union of subspaces, i.e., the assumption that each data sample can be represented as a linear combination of other samples in the same subspace.
The deep subspace clustering (DSC) network \cite{ji2017deep} is a deep auto-encoder based subspace clustering model to address the case of non-linear subspaces.
The authors of DSC introduced the self-expressive layer to integrate self-expressiveness property into a deep neural network.
This deep learning-based method was shown to outperform the other state-of-the-art subspace clustering methods significantly.
However, utilization of DSC is restricted to "shallow" models because the self-expressive layer requires a massive number of parameters.

In this paper, we propose a deep neural network to improve efficiency of self-expressiveness, which is termed \textbf{D}eep \textbf{C}losed-\textbf{F}orm \textbf{S}ubspace \textbf{C}lustering (\textbf{DCFSC}).
It consists of a closed form solution of the self-expressive layer motivated by \emph{EASE$^R$} model \cite{steck2019embarrassingly}, which showed that a similar Top-$N$ recommendation problem could be solved in closed form by a method of Lagrange multipliers.
We modified the self-expressive layer from the parameterized fully-connected layer to a closed-form solution.
In contrast to DSC, since the proposed self-expressive layer does not have any parameters for optimization, it is both memory-efficient and methodologically simple.
To the best of our knowledge, this is the very first attempt that proposes to use closed-form solution to self-expressive layer.
Furthermore, our model can use deeper neural networks for getting richer representation on subspace clustering.

We extensively evaluated our model on face clustering, using the Extended Yale B and ORL dataset for a small case, and on general object clustering, using COIL100 for a large case.
Our experiments showed that DCFSC achieved comparable performance only using 0.25\% \textasciitilde{} 0.44\% parameters of DSC on a small case, and the state-of-the-art result on a large case.

\section{Related Works}
Subspace clustering problem usually is divided by two subproblems.
The first subproblem is finding an affinity matrix from data.
The second subproblem is clustering data points using the affinity matrix via normalized cuts \cite{shi2000normalized} or spectral clustering \cite{ng2002spectral}.
Since there are already many good articles \cite{parsons2004subspace, kriegel2009clustering, vidal2011subspace} that dealt comprehensively with classic subspace clustering methods, here we only deal with recent works on subspace clustering related with deep representation learning.

Several works \cite{peng2016deep, peng2018structured, chen2018subspace} proposed a type of the methodology that representations learned by auto-encoder were forced to follow a specific conventional prior structure related with self-expression, e.g., Sparse Subspace Clustering (SSC) \cite{elhamifar2013sparse} and Low-rank Representation (LRR) \cite{liu2012robust}. 
\cite{li2017projective} proposed deep-encoder based row space recovery methodology to make conventional low-rank subspace clustering scalable and fast. 
\cite{peng2017cascade} simultaneously learned a compact representation using a neural network, and
\cite{zhang2018scalable} proposed combined methodology of a variant of K-subspace clustering \cite{balzano2012k} and deep auto-encoder to bypass the steps of constructing an affinity matrix and performing spectral clustering. 

On the other hand, \cite{ji2017deep} firstly introduced deep subspace clustering network.
The biggest contribution of \cite{ji2017deep} was that they firstly designed the self-expressive layer and corresponding loss function which models self-expressiveness property of data into deep auto-encoder.
Since DSC showed great performance on various benchmarks, there have been many subsequent studies \cite{zhou2018deep, zhang2019selfsupervised, zhou2019deep, zhang2019neural} that tried to improve the DSC in several aspects.
Deep adversarial subspace clustering \cite{zhou2018deep} exploited GAN-like adversarial learning framework to supervise sample representation learning and subspace clustering.
\cite{zhang2019selfsupervised} proposed a dual self-supervision framework which exploited the output of spectral clustering to supervise the training of the feature learning module and the self-expression module. 
\cite{zhou2019deep} introduced a new type of loss called distribution consistency loss to guide learning of distribution-preserving latent representation.
\cite{zhang2019neural} re-formulated subspace clustering as a classification problem, which in turn removed the spectral clustering step from the computations.

Despite the effectiveness and impact of DSC, the disadvantages of DSC have also been pointed out in some studies \cite{zhang2018scalable, zhang2019neural}.
The main disadvantage of DSC, which was commonly pointed out in these works, is that the memory footprint for training DSC is too high to access the subspace clustering problem for the large-scale dataset.
This memory problem is caused by two factors of DSC.
First, the self-expressive layer consists of $N^2$ parameters for $N$-size dataset. Second, the process of clustering data points from the affinity matrix has a quiet high memory requirement.
Neural Collaborative Subspace Clustering \cite{zhang2019neural} tried to solve the latter, but did not face the former problem.
Therefore, this paper is the primary work to solve the memory requirement problem of DSC's self-expressive layer.

\section{Proposed Framework}
\subsection{Deep Subspace Clustering}
Here, we firstly give a brief introduction on deep subspace clustering \cite{ji2017deep}, which is one of key papers in this work.
The core of DSC is joint training of deep auto-encoder and self-expressive layer.
Let $AE_{\Theta_{ae}}: \mathbb{R}^{N \times D} \to \mathbb{R}^{N \times D}$ denote the auto-encoder, which is parameterized with $\Theta_{ae}$.\footnote{In the rest of the paper, when there is no confusion, the subscript that represent the learnable parameters is omitted sometimes for simplicity of notation. This applies not only to $AE_{\Theta_{ae}}$, but also to all parameterized functions.}
$AE$ consists of two parts of feed-forward functions, a encoder $Enc_{\Theta_{enc}}: \mathbb{R}^{N \times D} \to \mathbb{R}^{N \times d}$ and a decoder $Dec_{\Theta_{dec}}: \mathbb{R}^{N \times d} \to \mathbb{R}^{N \times D}$.
$Enc$ and $Dec$ are parameterized with $\Theta_{enc}$ and $\Theta_{dec}$, respectively.
Let matrix $X \in \mathbb{R}^{N \times D}$ represent a $N$-size whole dataset. Each row of $X$ refers to each $D$-dimensional data point.
Standard auto-encoder is trained to optimize (L2) reconstruction error $L(X; \Theta_{ae})$:
\begin{equation}
\label{eq:ae}
L(X; \Theta_{ae}) = \norm{X - AE(X)}^2_F = \norm{X - Dec(Enc(X))}^2_F .
\end{equation}
However, training with Equation \ref{eq:ae}, the latent representation $Enc(X)$ is not guaranteed to have any beneficial property for subspace clustering.

For the guarantee, DSC utilizes self-expressiveness property of data drawn from union of linear subspaces \cite{rao2008motion, elhamifar2009sparse, elhamifar2013sparse}.
Self-expressiveness property of set of points is that there exists a matrix $C \in \mathbb{R}^{N \times N}$ which satisfies $X = C X$ if each row data of $X$ are drawn from one of the multiple linear subspaces.
$C$ is called self-representation coefficient matrix.
If each subspace is independent with other subspace, self-representation coefficient matrix $C$ has a block-diagonal structure \cite{ji2014efficient}.
With matrix norm constraint on $C$, finding optimal $C$ under these two assumptions is formulated as the following:
\begin{align}
\label{eq:self-expressiveness}
\min_{C} \norm{C}_p \text{\quad s.t.\quad} X = C X,\text{\:} \text{diag}(C) = 0.
\end{align}
Usually, complex high-dimensional data points in original data space itself do not satisfy self-expressiveness property so appropriate $C$ cannot be found.

Instead of building assumption of self-expressiveness on data space, DSC enforces latent space of data $Enc(X)$ to satisfy self-expressiveness property while training deep auto-encoder.
Self-representation coefficient matrix is instantiated as trainable parameters of self-expressive layer $SEL_{\Theta_{sel}}: \mathbb{R}^{N \times d} \to \mathbb{R}^{N \times d}$.
$\Theta_{sel} \in \mathbb{R}^{N \times N}$ denotes parameters of self-expressive layer.
Mapping by self-expressive layer is simply expressed as linear mapping among input data i.e.\ $SEL(Y) = \Theta_{sel} Y$ where $Y \in \mathbb{R}^{N \times d}$.
Under DSC framework, $AE(X)$ is defined as $Dec(SEL(Enc(X)))$. $\Theta_{ae}=\{\Theta_{enc}, \Theta_{dec}\}$ and $\Theta_{sel}$ are jointly optimized with constraints and regularization derived from the self-expressiveness property (Equation \ref{eq:self-expressiveness}):
\begin{align}
\label{eq:dsc}
L(X; \Theta_{ae}, \Theta_{sel}) = &\norm{X - Dec(SEL(Enc(X)))}^2_F \\ &+ \lambda_1 \norm{\Theta_{sel}}_p \nonumber \\ &+ \frac{\lambda_2}{2} \norm{Enc(X) - SEL(Enc(X))}^2_F \nonumber \\
\text{s.t.\quad} & \text{diag}(\Theta_{sel}) = 0, \nonumber
\end{align}
where $p = 1 $ or $p = 2$ in \cite{ji2017deep}. In this work, we only consider $p=2$ case.
This is because using L-2 norm makes optimization free from the diagonal constraint \cite{ji2014efficient} and usually shows better performance than using L-1 norm.
\begin{algorithm}[t]
    \caption{Deep Subspace Clustering \cite{ji2017deep}}
    \label{algo:dsc}
    \hspace*{\algorithmicindent} \textbf{Input}: Data Matrix $X$,\\
    \hspace*{\algorithmicindent}\phantom{\textbf{Input}: } Encoder with Pre-trained Parameters $Enc_{\Theta{enc}}$,\\
    \hspace*{\algorithmicindent}\phantom{\textbf{Input}: } {\color{Magenta}Parameters of Self-expressive Layer $\Theta{sel}$},\\
    \hspace*{\algorithmicindent}\phantom{\textbf{Input}: } Decoder with Pre-trained Parameters $Dec_{\Theta{dec}}$,\\
    \hspace*{\algorithmicindent}\phantom{\textbf{Input}: } {\color{Magenta}Hyper-parameters for Loss Weights $\lambda_1, \lambda_2$},\\
    \hspace*{\algorithmicindent}\phantom{\textbf{Input}: } Number of Training Iteration $EndStep$\\
    \hspace*{\algorithmicindent} \textbf{Output} ${\Tilde{\Theta}_{sel}}$
\begin{algorithmic}[1]

\State $\Theta \gets \{\Theta_{enc}, {\color{Magenta}\Theta{sel}}, \Theta{dec}\}$
\State $n\_iter \gets 0$
\While {$n\_iter < EndStep$}
  \State $X_{latent} \gets Enc_{\Theta_{enc}}(X)$
  \State {\color{Magenta}${X_{latent}}^{\prime} \gets \Theta{sel}\text{ }X_{latent}$}
  \State $X_{recon} \gets Dec_{\Theta_{dec}}({X_{latent}}^{\prime})$
  \State {\color{Magenta}Update $\Theta$ to minimize Equation \ref{eq:dsc}}
  \State $n\_iter \gets n\_iter + 1$
\EndWhile
\State ${\Tilde{\Theta}_{sel}} \gets \Theta_{sel}$\\
\Return ${\Tilde{\Theta}_{sel}}$
\end{algorithmic}
\end{algorithm}
Algorithm \ref{algo:dsc} shows the whole training scheme of DSC.

After the network is trained, parameters of self-expressive layer $\Theta_{sel}$ is used for constructing affinity matrix $A \in \mathbb{R}^{N \times N}$.
This affinity matrix is then used for spectral clustering \cite{ng2002spectral} to yield the final data clustering result.
For building affinity matrix from the parameters of the self-expressive layer, the official implementation of DSC utilizes sparse subspace clustering (SSC) algorithm \cite{elhamifar2013sparse}.
To cluster data points from the affinity matrix, spectral clustering method \cite{ng2002spectral} is used.
\subsection{Closed Form Solution of Self-Expressive Layer}
In this section, we consider the following optimization problem:
\begin{align}
\label{eq:sae}
\min\limits_{\Theta_{sel}} \norm{X^{\prime} - \Theta_{sel}X^{\prime}}^2_F + \lambda \norm{\Theta_{sel}}^2_F \text{\quad s.t.\quad} & \text{diag}(\Theta_{sel}) = 0.
\end{align}
Problem \ref{eq:sae} is the partial problem of DSC's objective when $X^{\prime}$ is defined as $Enc_{\Theta_{enc}}(X)$.
Specifically, Problem \ref{eq:sae} is the problem excluding auto-encoding loss term (Equation \ref{eq:ae}) from the minimization problem of the DSC's objective function (Equation \ref{eq:dsc}).

The optimization problem \ref{eq:sae} is usually dealt in several works \cite{ning2011slim, ning2012sparse, cheng2014lorslim, steck2019embarrassingly} in field of Top-$N$ recommendation problem.
Recently, \cite{steck2019embarrassingly} showed that this problem could be simply solved in closed form by method of Lagrange multipliers.
Motivated from it, our main approach to optimize Equation \ref{eq:dsc} is to adopt this closed-form solution of Problem \ref{eq:sae} and minimize only Equation \ref{eq:ae}, instead of minimizing Equation \ref{eq:dsc} by first-order methods.

Following derivation in Section 3.1 of \cite{steck2019embarrassingly}, the closed-form solution of Equation \ref{eq:sae} is given as the following:
\begin{align}
\label{eq:closed-form}
B = I - P \cdot \text{diagMat}(\overrightarrow{1} \oslash \text{diag}(P)),
\end{align}
where $P = (XX^T + \lambda I)^{-1}$. In Equation \ref{eq:closed-form}, $\text{diagMat}(\cdot)$, $\overrightarrow{1}$ $\oslash$, and $\text{diag}(\cdot)$ denote operation converting vector to diagonal matrix, a vector of ones, Hadamard division of matrices, and operation converting diagonal matrix to vector, in order. With reconfiguration of Equation \ref{eq:closed-form}, the solution can become more computationally efficient form:
\begin{equation}
B{ij} = \left\{
  \begin{array}{ll}
     0& {\rm \text{if}}\,\,\, i=j\\
     -\frac{P_{ij}}{P_{ii}}&{\rm \text{otherwise}.}
  \end{array}
\right .
\label{eq_w2}
\end{equation}
\subsection{Deep Closed-Form Subspace Clustering}
\begin{algorithm}[t]
    \caption{Deep Closed-Form Subspace Clustering}
    \label{algo:dcfsc}
    \hspace*{\algorithmicindent} \textbf{Input}: Data Matrix $X$,\\
    \hspace*{\algorithmicindent}\phantom{\textbf{Input}: } Encoder with Parameters $Enc_{\Theta{enc}}$,\\
    \hspace*{\algorithmicindent}\phantom{\textbf{Input}: } Decoder with Parameters $Dec_{\Theta{dec}}$,\\
    \hspace*{\algorithmicindent}\phantom{\textbf{Input}: } {\color{MidnightBlue}Matrix Regularization Parameter $\lambda$},\\
    \hspace*{\algorithmicindent}\phantom{\textbf{Input}: } Number of Training Iteration $EndStep$\\
    \hspace*{\algorithmicindent} \textbf{Output} Self-representation Coefficient Matrix $B$
\begin{algorithmic}[1]

\State $\Theta \gets \{\Theta_{enc}, \Theta{dec}\}$
\State $n\_iter \gets 0$
\While {$n\_iter < EndStep$}
  \State $X_{latent} \gets Enc_{\Theta_{enc}}(X)$
  \State {\color{MidnightBlue}$P \gets \text{compute\_p}(X_{latent}, \lambda)$}
  \State {\color{MidnightBlue}$B \gets \text{compute\_b(P)}$}
  \State $\bar{B} \gets \text{Stop\_Gradients}
  (B)$
  \State {\color{MidnightBlue}${X_{latent}}^{\prime} \gets \bar{B}\text{ }X_{latent}$}
  \State $X_{recon} \gets Dec_{\Theta_{dec}}({X_{latent}}^{\prime})$
  \State {\color{MidnightBlue}Update $\Theta$ to minimize Equation \ref{eq:ae}}
  \State $n\_iter \gets n\_iter + 1$
\EndWhile
\Return $B$
\end{algorithmic}
\end{algorithm}
\begin{lstfloat}
\begin{lstlisting}[language=Python, caption=Implementation of $compute\_p$ and $compute\_b$ on Tensorflow., label={code:imp}][t]
import tensorflow as tf
 
def compute_p(encoded, coef_lambda):
    # In: encoded (Tensor with shape [N, d])
    # In: coef_lambda (float)
    # Out: Tensor with shape [N, N]
    encoded_t = tf.transpose(encoded)
    gram_matrix = tf.matmul(encoded, encoded_t)
    identity = tf.eye(encoded.shape[0])
    gram_matrix += coef_lambda * identity
    p = tf.linalg.inv(gram_matrix)
    return p

def compute_b(p):
    # In: p (Tensor with shape [N, N])
    # Out: Tensor with shape [N, N]
    diag_p = tf.linalg.diag_part(p)
    b = p / (- diag_p[:, tf.newaxis])
    
    zeros = tf.zeros(b.shape[0:-1])
    b = tf.linalg.set_diag(b, zeros)
    return b
\end{lstlisting}
\end{lstfloat}

Our DCFSC is a variant of DSC with closed form solution of self-expressive layer.
Algorithm \ref{algo:dcfsc} describes how training procedure of DCFSC works.
The main differences between Algorithm \ref{algo:dsc} and Algorithm \ref{algo:dcfsc} are indicated by magenta and blue, respectively.
Two core steps of DCFSC,\ $\text{compute\_p}(\cdot, \cdot)$ and $\text{compute\_b}(\cdot)$, are directly matched with Equation \ref{eq:closed-form}.
Listing \ref{code:imp} is Tensorflow \cite{abadi2016tensorflow} implementation for $\text{compute\_p}(\cdot, \cdot)$ and $\text{compute\_b}(\cdot)$.
As the readers can see, DCFSC is easy to implement as much as DSC.

Compared with DSC (Algorithm \ref{algo:dsc}), DCFSC does not retain $N \times N$-size parameters for self-representation coefficient matrix so does not need to optimize them.
Moreover, there is no need to compute gradient over $B$ or $P$ because closed form solution for self-representation coefficient matrix is directly derived from $X_{latent}$ only via forward pass.

In case of small dataset such as ORL ($N=400$) and Extended Yale B ($N=2,432$), it results in little benefit over the existing DSC method.
However, if size of of dataset is relatively large like COIL-100 ($N=7,200$), our approach has a significant benefit on memory efficiency.
On large datasets, advantages of DCFSC over DSC enable us to use deeper architecture to get better latent representations for subspace clustering.
In contrast to DCFSC, DSC only allows shallow models to be used because of memory issue related with self-representation coefficient matrix.
\section{Experiments}
\label{sec:experiments}
\paragraph{Compared Methods and Performance Metric}
For performance comparison among several subspace clustering methods, we adopted list of methods and benchmark results from the previous works \cite{peng2016deep, ji2017deep}: Low Rank Representation (LRR) \cite{liu2012robust}, Low Rank Subspace Clustering (LRSC) \cite{vidal2014low}, Sparse Subspace Clustering (SSC) \cite{elhamifar2013sparse}, Kernel Sparse Subspace Clustering (KSSC) \cite{patel2014kernel}, SSC by Orthogonal Matching Pursuit (SSCOMP) \cite{you2016scalable}, Efficient Dense Subspace Clustering (EDSC) \cite{ji2014efficient}, SSC with the pre-trained convolutional auto-encoder features (AE+SSC), EDSC with the pre-trained convolutional auto-encoder features (AE+EDSC), and Deep Subspace Clustering (DSC) \cite{ji2017deep}.
Since the performances of DSC with L2 regularization were reported to be consistently better than those of L1 regularization, only performances of L2-regularized version of DSC were reported here. We also used the clustering error rate as metric for evaluating performance of each subspace clustering method, as same with \cite{ji2017deep}.
We collected benchmark results of various methods from the DSC paper.

\paragraph{Design of Experiments}
We separated experiments into two cases by size of dataset: small $N$ cases (Section \ref{sec:face}) and large $N$ case (Section \ref{sec:object}).

The design of small $N$ case experiments was to show performance of DCFSC under the very same settings of DSC paper.
The only difference were the algorithm part of DSC and DCFSC.
The other settings of experiments (e.g., model architecture, training procedure, and evaluation protocol) were same with ones of the original DSC paper.
In terms of performance, it might be quiet unfavorable and unfair for DCFSC because DCFSC has much smaller model parameters than DSC in same architecture setting.
Thus, design of these experiments was intended to answer how well DCFSC worked in exactly the same settings as DSC in its paper, regardless of superior point of DCFSC on memory efficiency.
In the small $N$ cases, Extended Yale B dataset \cite{lee2005acquiring} and ORL dataset \cite{samaria1994parameterisation} were used.

In contrast to small $N$ cases, the experiment on large $N$ case was designed to verify performance with full use of DCFSC's memory efficiency.
In the experiment, convolutional auto-encoder architecture, which was deeper than that used in the work of DSC, was used for implementation of our DCFSC.
Note that this deeper architecture was quiet computationally intractable under the DSC method.
Thus, this experiment was intended to show our DCFSC's computational efficiency and the followed possibility of stronger representation learning.
COIL-100 dataset \cite{nene1996columbia} was used for large $N$ case.

\paragraph{System Environment}
Implementation of DCFSC for experiments was done with minimum modification of public implementation of the DSC paper.\footnote{\url{https://github.com/panji1990/Deep-subspace-clustering-networks}}
The Python version used was 3.5.2, and the Tensorflow version was 1.8.0.
In addition, a single NVIDIA TESLA V100 GPU with 40 Intel Xeon E5-2698 CPUs were used for the experiment, and the CUDA and CuDNN version used were 9.0 and 7.1.4, respectively.

\subsection{Small $N$ Case: E-YaleB and ORL}
\label{sec:face}
\begin{table*}[t]
\centering
\begin{tabular}{lccccccc}
\multicolumn{1}{l|}{Layers} & \multicolumn{1}{l}{Encoder-1} & \multicolumn{1}{l}{Encoder-2} & \multicolumn{1}{l}{Encoder-3} & \multicolumn{1}{l}{Self-Expressive} & \multicolumn{1}{l}{Decoder-1} & \multicolumn{1}{l}{Decoder-2} & \multicolumn{1}{l}{Decoder-3} \\ \hline
\multicolumn{8}{c}{Deep Subspace Clustering (Total \# of Parameters: 5,929,615)} \\ \hline
\multicolumn{1}{l|}{Kernel Size} & 5 $\times$ 5 & 3 $\times$ 3 & 3 $\times$ 3 & - & 3 $\times$ 3 & 3 $\times$ 3 & 5 $\times$ 5 \\
\multicolumn{1}{l|}{\# of Channels} & 10 & 20 & 30 & - & 30 & 20 & 10 \\
\multicolumn{1}{l|}{\# of Parameters} & 260 & 1,820 & 5,430 & 5,914,624 & 5,420 & 1,810 & 251 \\ \hline
\multicolumn{8}{c}{Deep Closed-Form Subspace Clustering (Total \# of Parameters: 14,991)} \\ \hline
\multicolumn{1}{l|}{Kernel Size} & 5 $\times$ 5 & 3 $\times$ 3 & 3 $\times$ 3 & - & 3 $\times$ 3 & 3 $\times$ 3 & 5 $\times$ 5 \\
\multicolumn{1}{l|}{\# of Channels} & 10 & 20 & 30 & - & 30 & 20 & 10 \\
\multicolumn{1}{l|}{\# of Parameters} & 260 & 1,820 & 5,430 & 0 & 5,420 & 1,810 & 251
\end{tabular}
\caption{Comparison on Network settings for Extended Yale B. DCFSC has only model parameters of $\frac{14,991}{5,929,615} \times 100 \% = 0.25 \% $ as compared to DSC.}
\label{tab:eyale_arch}
\end{table*}
\begin{table*}[t]
\centering
\begin{tabular}{lccccccc}
\multicolumn{1}{l|}{Layers} & \multicolumn{1}{l}{Encoder-1} & \multicolumn{1}{l}{Encoder-2} & \multicolumn{1}{l}{Encoder-3} & \multicolumn{1}{l}{Self-Expressive} & \multicolumn{1}{l}{Decoder-1} & \multicolumn{1}{l}{Decoder-2} & \multicolumn{1}{l}{Decoder-3} \\ \hline
\multicolumn{8}{c}{Deep Subspace Clustering (Total \# of Parameters: 160,702)} \\ \hline
\multicolumn{1}{l|}{Kernel Size} & 5 $\times$ 5 & 3 $\times$ 3 & 3 $\times$ 3 & - & 3 $\times$ 3 & 3 $\times$ 3 & 5 $\times$ 5 \\
\multicolumn{1}{l|}{\# of Channels} & 5 & 3 & 3 & - & 3 & 3 & 5 \\
\multicolumn{1}{l|}{\# of Parameters} & 130 & 138 & 84 & 160,000 & 84 & 140 & 126 \\ \hline
\multicolumn{8}{c}{Deep Closed-Form Subspace Clustering (Total \# of Parameters: 702)} \\ \hline
\multicolumn{1}{l|}{Kernel Size} & 5 $\times$ 5 & 3 $\times$ 3 & 3 $\times$ 3 & - & 3 $\times$ 3 & 3 $\times$ 3 & 5 $\times$ 5 \\
\multicolumn{1}{l|}{\# of Channels} & 5 & 3 & 3 & - & 3 & 3 & 5 \\
\multicolumn{1}{l|}{\# of Parameters} & 130 & 138 & 84 & 0 & 84 & 140 & 126
\end{tabular}
\caption{Comparison on Network settings for ORL. DCFSC has only model parameters of $\frac{702}{160,702} \times 100 \% = 0.44 \% $ as compared to DSC.}
\label{tab:orl_arch}
\end{table*}
\begin{table*}[t]
\centering
\vspace{1cm}
\begin{tabular}{lcccccccccc}
\multicolumn{1}{c|}{Method} & LRR & LRSC & SSC & \begin{tabular}[c]{@{}c@{}}AE+\\ SSC\end{tabular} & KSSC & \begin{tabular}[c]{@{}c@{}}SSC-\\ OMP\end{tabular} & EDSC & \begin{tabular}[c]{@{}c@{}}AE+\\ EDSC\end{tabular} & DSC & DCFSC \\ \hline
\multicolumn{11}{c}{10 subjects} \\ \hline
\multicolumn{1}{l|}{Mean} & 22.22 & 30.95 & 10.22 & 17.06 & 14.49 & 12.08 & 5.64 & \underline{5.46} & \textbf{1.59} & 5.72 \\
\multicolumn{1}{l|}{Median} & 23.49 & 29.38 & 11.09 & 17.75 & 15.78 & 8.28 & \underline{5.47} & 6.09 & \textbf{1.25} & 5.63 \\ \hline
\multicolumn{11}{c}{15 subjects} \\ \hline
\multicolumn{1}{l|}{Mean} & 23.22 & 31.47 & 13.13 & 18.65 & 16.22 & 14.05 & 7.63 & 6.70 & \textbf{1.69} & \underline{5.33} \\
\multicolumn{1}{l|}{Median} & 23.49 & 31.64 & 13.40 & 17.76 & 17.34 & 14.69 & 6.41 & 5.52 & \textbf{1.72} & \underline{5.10} \\ \hline
\multicolumn{11}{c}{20 subjects} \\ \hline
\multicolumn{1}{l|}{Mean} & 30.23 & 28.76 & 19.75 & 18.23 & 16.55 & 15.16 & 9.30 & 7.67 & \textbf{1.73} & \underline{4.93} \\
\multicolumn{1}{l|}{Median} & 29.30 & 28.91 & 21.17 & 16.80 & 17.34 & 15.23 & 10.31 & 6.56 & \textbf{1.80} & \underline{4.92} \\ \hline
\multicolumn{11}{c}{25 subjects} \\ \hline
\multicolumn{1}{l|}{Mean} & 27.92 & 27.81 & 26.22 & 18.72 & 18.56 & 18.89 & 10.67 & 10.27 & \textbf{1.75} & \underline{4.92} \\
\multicolumn{1}{l|}{Median} & 28.13 & 26.81 & 26.66 & 17.88 & 18.03 & 18.53 & 10.84 & 10.22 & \textbf{1.81} & \underline{5.00} \\ \hline
\multicolumn{11}{c}{30 subjects} \\ \hline
\multicolumn{1}{l|}{Mean} & 37.98 & 30.64 & 28.76 & 19.99 & 20.49 & 20.75 & 11.24 & 11.56 & \textbf{2.07} & \underline{5.35} \\
\multicolumn{1}{l|}{Median} & 36.82 & 30.31 & 28.59 & 20.00 & 20.94 & 20.52 & 11.09 & 10.36 & \textbf{2.19} & \underline{5.52} \\ \hline
\multicolumn{11}{c}{35 subjects} \\ \hline
\multicolumn{1}{l|}{Mean} & 41.85 & 31.35 & 28.55 & 22.13 & 26.07 & 20.29 & 13.10 & 13.28 & \textbf{2.65} & \underline{5.93} \\
\multicolumn{1}{l|}{Median} & 41.81 & 31.74 & 29.04 & 21.74 & 25.92 & 20.18 & 13.10 & 13.21 & \textbf{2.64} & \underline{5.96} \\ \hline
\multicolumn{11}{c}{38 subjects} \\ \hline
 & 34.87 & 29.89 & 27.51 & 25.33 & 27.75 & 24.71 & 11.64 & 12.66 & \textbf{2.67} & \underline{6.13}
\end{tabular}
\caption{Clustering error (in \%) on Extended Yale B. The lower the better. Lower is better. The \textbf{bold} and \underline{underlined} text refer to the 1st and 2nd ranked score, respectively.}
\label{tab:eyale_bench}
\end{table*}

\begin{figure*}[t]
     \centering
     \vspace{1cm}
     \begin{subfigure}[b]{0.5\textwidth}
         \centering
         \includegraphics[width=\textwidth]{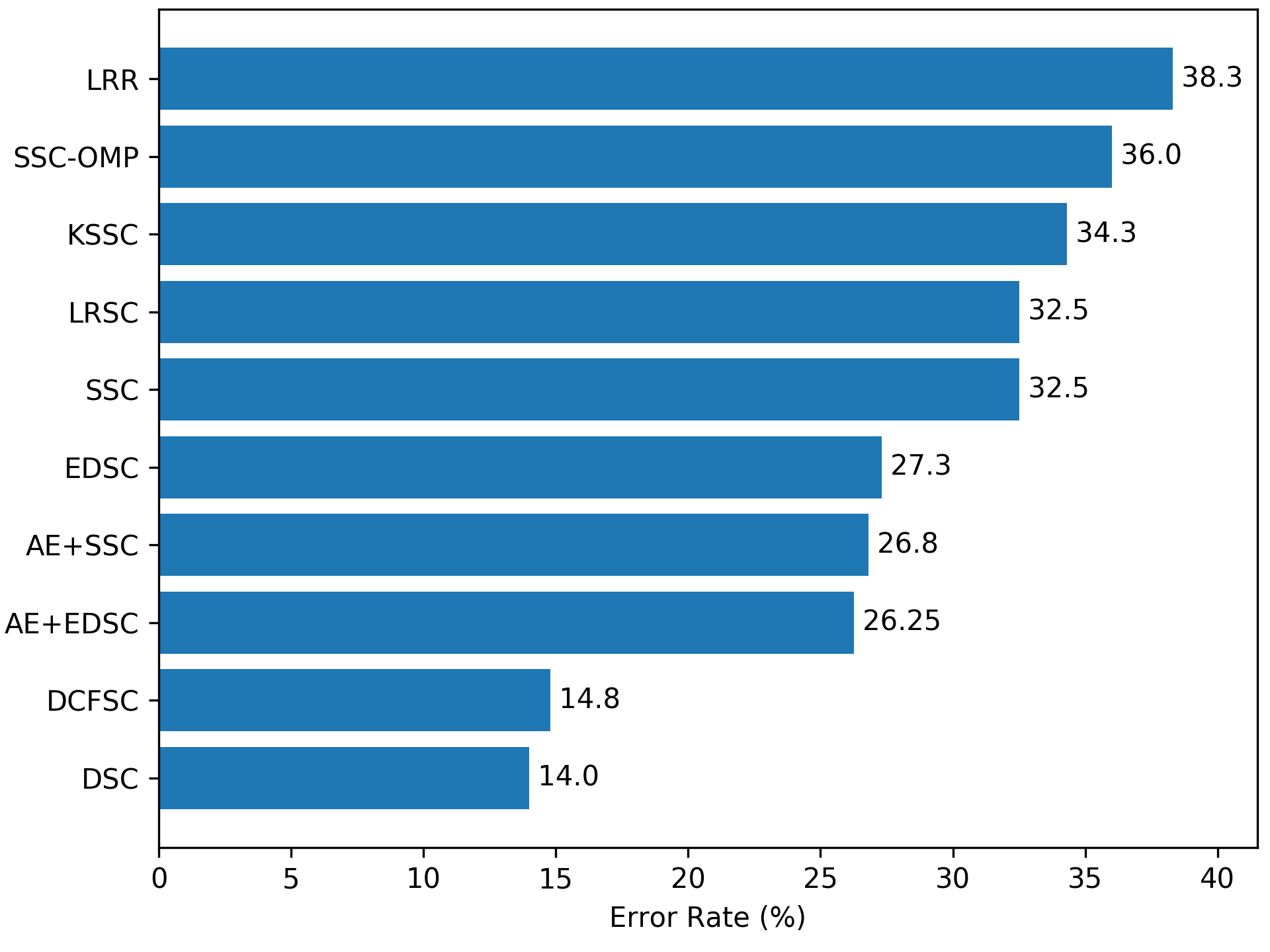}
         \caption{ORL}
         \label{fig:orl_bench}
     \end{subfigure}%
     \begin{subfigure}[b]{0.5\textwidth}
         \centering
         \includegraphics[width=\textwidth]{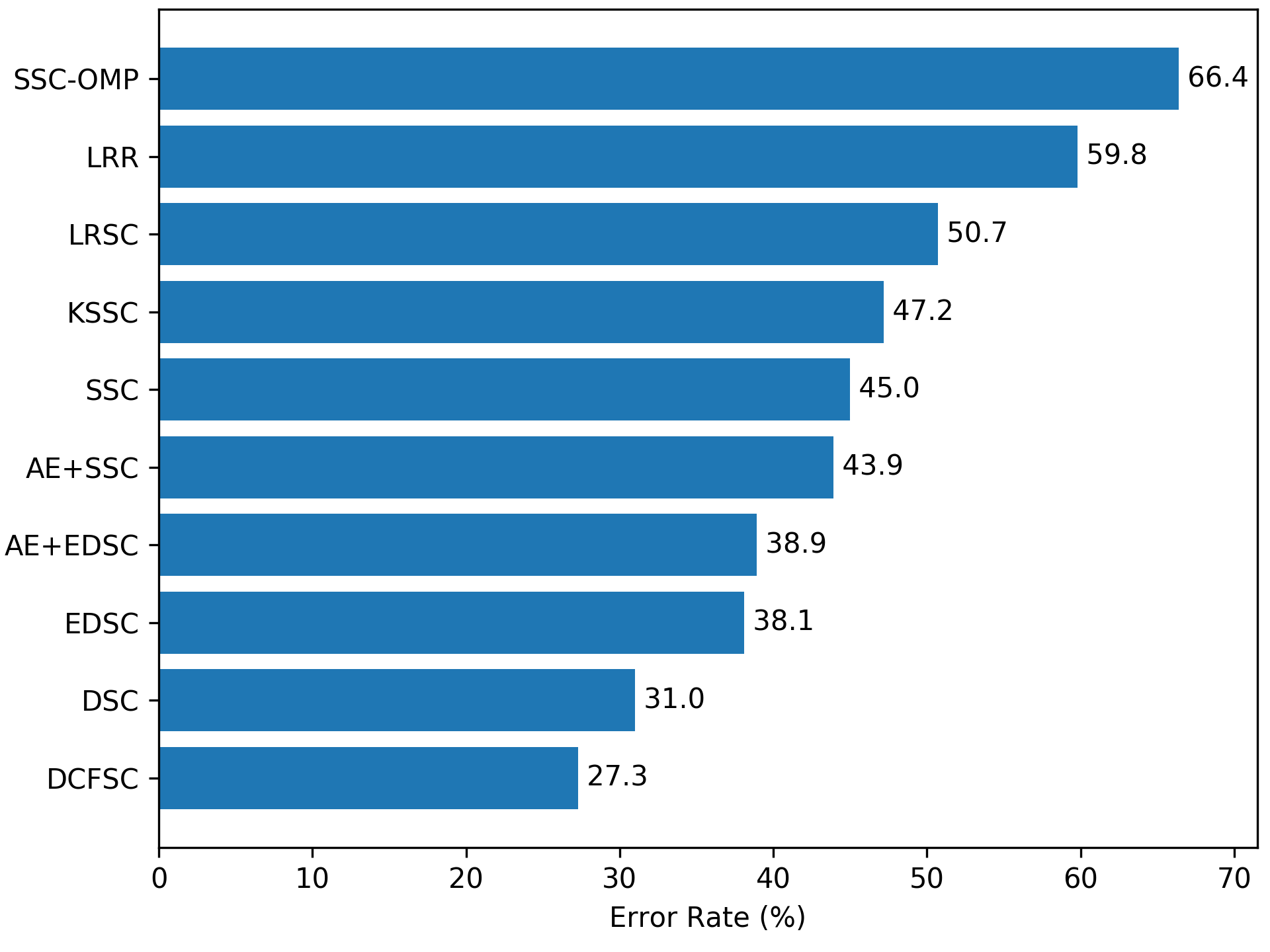}
         \caption{COIL-100}
         \label{fig:coil_bench}
     \end{subfigure}
    \caption{Subspace clustering error on ORL and COIL100. Methods are sorted in descending order of error. Lower is better. For both benchmarks of DCFSC, mean of ten trails is reported.}
    \label{fig:bench}
\end{figure*}
\paragraph{Data Description}
Both Extended Yale B (E-YaleB) \cite{lee2005acquiring} and ORL \cite{samaria1994parameterisation} are face databases.
Tuples representing number of classes, number of images per class, total number of images, and size of images on E-YaleB and ORL dataset are $(38, 64, 2432, 192 \times 168)$ and $(40, 10, 400, 112 \times 92)$, respectively.
The main difficulty of E-YaleB is known as extreme illumination, whereas the difficulty of ORL is known as deformation and various pose.
Like experiment setting of the DSC paper \cite{ji2017deep}, images of E-YaleB were resized to $48 \times 48$, and images of ORL were resized to $32 \times 32$.

\paragraph{Experiment Settings}
For small $N$ cases, we used the almost same  neural architecture used in DSC with little modification.
The only difference was that we removed the self-expressive layers from the DSC network and changed the training algorithm from Algorithm \ref{algo:dsc} to Algorithm \ref{algo:dcfsc}.
Table \ref{tab:eyale_arch} and Table \ref{tab:orl_arch} show overall comparisons of number of parameters between DSC and DCFSC for two small $N$ cases experiments.
Note that we did not extensively search for any other optimal training hyper-parameter or neural architecture for DCFSC.

All other settings of experiments for E-YaleB dataset and ORL dataset were same with experiments in the DSC paper.
To measure the robustness of the DCFSC model for various numbers of clusters, we measured performance on several $K$ subjects in the E-YaleB dataset.
Here, number of clusters $K$ was \{10, 15, 20, 25, 30, 35, 38\} and each subject was set to have 64 face images.
For ORL dataset, number of clusters was set to 40, just like the original subject number of the ORL dataset.
For both E-YaleB and ORL, learning rate and matrix regularization parameter $\lambda$ in the DCFSC are set as $0.001$ and $5e^5$, respectively.
The model weights in the DCFSC were initialized to the pre-trained weights used in the DSC experiments.
For fine-tuning stage, the DCFSC model was trained by $50 + 25K$ epochs for each $K$ in E-YaleB dataset and by $700$ epochs in ORL dataset.

\begin{table*}[ht]
\begin{tabular}{lccccccccccc}
\hline
\multicolumn{12}{c}{Deep Subspace Clustering (Total \# of Parameters: 51,842,600)} \\ \hline
\multicolumn{1}{l|}{Layers} & \multicolumn{4}{c}{Encoder-1} & \multicolumn{3}{c}{Self-Expressive} & \multicolumn{4}{c}{Decoder-1} \\ \hline
\multicolumn{1}{l|}{Kernel Size} & \multicolumn{4}{c}{5 $\times$ 5} & \multicolumn{3}{c}{-} & \multicolumn{4}{c}{5 $\times$ 5} \\
\multicolumn{1}{l|}{Stride Size} & \multicolumn{4}{c}{2} & \multicolumn{3}{c}{-} & \multicolumn{4}{c}{2} \\
\multicolumn{1}{l|}{\# of Channels} & \multicolumn{4}{c}{50} & \multicolumn{3}{c}{-} & \multicolumn{4}{c}{50} \\
\multicolumn{1}{l|}{\# of Parameters} & \multicolumn{4}{c}{1,300} & \multicolumn{3}{c}{51,840,000} & \multicolumn{4}{c}{1,300} \\
\multicolumn{12}{l}{} \\ \hline
\multicolumn{12}{c}{Deep Closed-Form Subspace Clustering (Total \# of Parameters: 81,913)} \\ \hline
\multicolumn{1}{l|}{Layers} & Enc-1 & Enc-2 & Enc-3 & Enc-4 & Enc-5 & \begin{tabular}[c]{@{}c@{}}Self-\\ Expressive\end{tabular} & Dec-1 & Dec-2 & Dec-3 & Dec-4 & Dec-5 \\ \hline
\multicolumn{1}{l|}{Kernel Size} & 5 $\times$ 5 & 3 $\times$ 3 & 3 $\times$ 3 & 3 $\times$ 3 & 1 $\times$ 1 & - & 1 $\times$ 1 & 3 $\times$ 3 & 3 $\times$ 3 & 3 $\times$ 3 & 5 $\times$ 5 \\
\multicolumn{1}{l|}{Stride Size} & 1 & 2 & 1 & 2 & 1 & - & 1 & 2 & 1 & 2 & 1 \\
\multicolumn{1}{l|}{\# of Channels} & 24 & 24 & 48 & 48 & 72 & - & 72 & 48 & 48 & 24 & 24 \\
\multicolumn{1}{l|}{\# of Parameters} & 696 & 5,280 & 10,560 & 20,928 & 3,528 & 0 & 3,648 & 20,928 & 10,464 & 5,280 & 601
\end{tabular}
\caption{Comparison on Network settings for COIL-100. DCFSC has only model parameters of $\frac{81,913}{51,842,600} \times 100 \% = 0.16 \% $ as compared to DSC. Note that on the architecture used in DCFSC batch normalization layers \cite{ioffe2015batch} were used except for the last layer of the encoder and decoder.}
\label{tab:coil100_arch}
\end{table*}

\paragraph{Results and Discussions}
In terms of the number of parameters, the DCFSC model had only 0.25\% and 0.44\%, compared with the DSC model.
The module that occupied most of the parameters in the DSC was the self-expressiveness layer.
DCFSC and DSC showed no significant difference in terms of memory requirements during training.
For instance, the amount of GPU memory required in training of DSC on ORL dataset was 1,022MB, whereas in DCFSC it was 942MB.
Therefore, in small $N$ cases, it is hard to say that DCFSC has a great advantage in learning procedure over DSC.

Benchmark results of small $N$ cases are shown in Table \ref{tab:eyale_bench} and Figure \ref{fig:orl_bench}.
For various number of clusters in the E-YaleB dataset, the DCFSC showed mean of clustering error rate of 6.13\%.
This was significantly worse than DSC's mean error rate (2.67\%), but it was much better than other hard baselines (11.64\% or higher).
On the ORL dataset, DCFSC resulted in an error rate of 14.8\%, which was slightly worse than 14.0\% of DSC (but not significantly).
In short, in the small $N$ cases, the DCFSC showed almost equal or worse performance than the DSC on nearly the same settings as the experiment in the DSC paper.
However, the performance of DCFSC was dominant over all other existing baselines except DSC.
The reason that DCFSC was inferior in performance to DSC might be that the neural architecture was not deep enough to yield the potential self-expression directly from data.
Therefore, there still is room for improvement in DCFSC performance, like searching for an optimal architecture or hyper-parameter.
These results, however, still show that convergence is experimentally guaranteed even in small $N$ cases.
\subsection{Large $N$ Case: COIL-100}
\label{sec:object}
\paragraph{Data Description}
For large $N$ case, COIL-100 dataset \cite{nene1996columbia}, which is a object database, was used to measure the performance of object clustering.
On COIL-100 dataset, number of classes, number of images per class, total number of images, and size of images are $100, 72, 7200$, and $128\times128$, respectively.
The main difficulties of dealing with COIL-100 dataset are known as deformation and rotation.
For consistency with previous studies \cite{cai2010graph, ji2017deep}, images of COIL-100 were resized to $32 \times 32$.

\paragraph{Experiment Settings}
The model architecture, used in the original DSC work \cite{ji2017deep} for COIL-100 dataset, was a very shallow auto-encoder structure consisting of one encoder layer, a self-expressive layer, and one decoder layer.
This was because the number of parameters that the self-expressive layer should retrain in case of large $N$ was too huge to adopt deeper neural architecture due to memory problems.
We used much deeper auto-encoder architecture to show that DCFSC could have tremendous advantages in this situation.
Table \ref{tab:coil100_arch} shows the difference between model architectures of DSC and DCFSC, used in the experiments.
Unlike DSC, five encoder layers and five decoder layers were used in DCFSC.

The number of clusters was set to 100, which was equal to the number of subjects in COIL-100.
Learning rate and matrix regularization parameters were set to $0.001$ and $10$, respectively.
While DSC had pre-trained model weights, the architecture of DCFSC in this experiment did not have such pre-trained model weights because new architecture was designed for training the DCFSC model.
Thus, in the large $N$ case experiment DCFSC was trained from scratch without pre-training.
Because the DCFSC model did not use pre-training weights, it was trained for 175 epochs, which was longer than 120 epochs used in the DSC.
All other experiment settings were same as in DSC's ones.

\paragraph{Results and Discussions}
Most of the model parameters of DSC, which were used for large $N$ case experiment, belonged to the self-expressive layer, and this tendency was much greater than in small $N$ cases.
This was because size of the self-expressive layer is proportional to the square of dataset size.
In reality, the DSC model for COIL-100 required huge amount of self-expressive layer parameters ($7,200 \times 7,200 = 51,840,000$).
By storing these parameters in double precision, simply maintaining these parameters required about 3.2 GB of memory space.
Furthermore, about 8.6GB of GPU memory was required to train the very shallow COIL-100 model presented in the DSC paper.
This drawback of DSC made it be not able to use deeper architecture.
Therefore, performance constraint of DSC on large dataset was practically inevitable.
Our DCFSC is free from these limitations of DSC model.
Number of parameters of the auto-encoder used in DCFSC was about 32 times of the DSC auto-encoder.
Despite using much deeper architecture than the one used in the DSC, the GPU memory requirement to train it was 10.8GB, which was only 26\% higher than the DSC architecture.
This means that the DCFSC can actually use a deeper architecture than the DSC.

Figure \ref{fig:coil_bench} shows several benchmark results on COIL-100 dataset.
Our DCFSC model showed a clustering error of 27.3\% and outperformed all other models including DSC, without pre-training.
These results reveal the performance advantage of DCFSC on larger dataset.
In addition, they implies that deep learning based subspace clustering still has room to benefit from learning richer latent representations through deeper architecture.
It is also noteworthy that performance of the DCFSC in this experiment was reported to be comparable with 26.6\%, which was reported in the study \cite{zhang2019selfsupervised} combining more sophisticated methodologies such as self-supervised learning with the DSC.
Since DCFSC is easy to combine with the more advanced models \cite{zhou2018deep, zhang2019selfsupervised, zhou2019deep, zhang2019neural} of DSC, there is possibility of further enhancing the performance of these modified models with deeper neural architecture.\footnote{It is also remarkable that in 'Deep Adversarial Subspace Clustering' paper \cite{zhou2018deep} the proposed model could not be used to try experiment in COIL-100, even with a very shallow auto-encoder model. It was also due to a memory shortage problem.}
\subsection{Effect of $\lambda$: Case of COIL-100}
We further investigated the effect of selection of matrix regularization parameter $\lambda$ on performance of DCFSC.
This is to see how sensitive DCFSC performance is to the choice of $\lambda$, or how robust it is.
For benchmarking, performances were reported by changing $\lambda$ from $1$ to $1e^6$ in multiples of 10 in the same settings as the COIL-100 dataset (Section \ref{sec:object}).

\begin{figure}[t]
     \centering
     \includegraphics[width=0.5\textwidth]{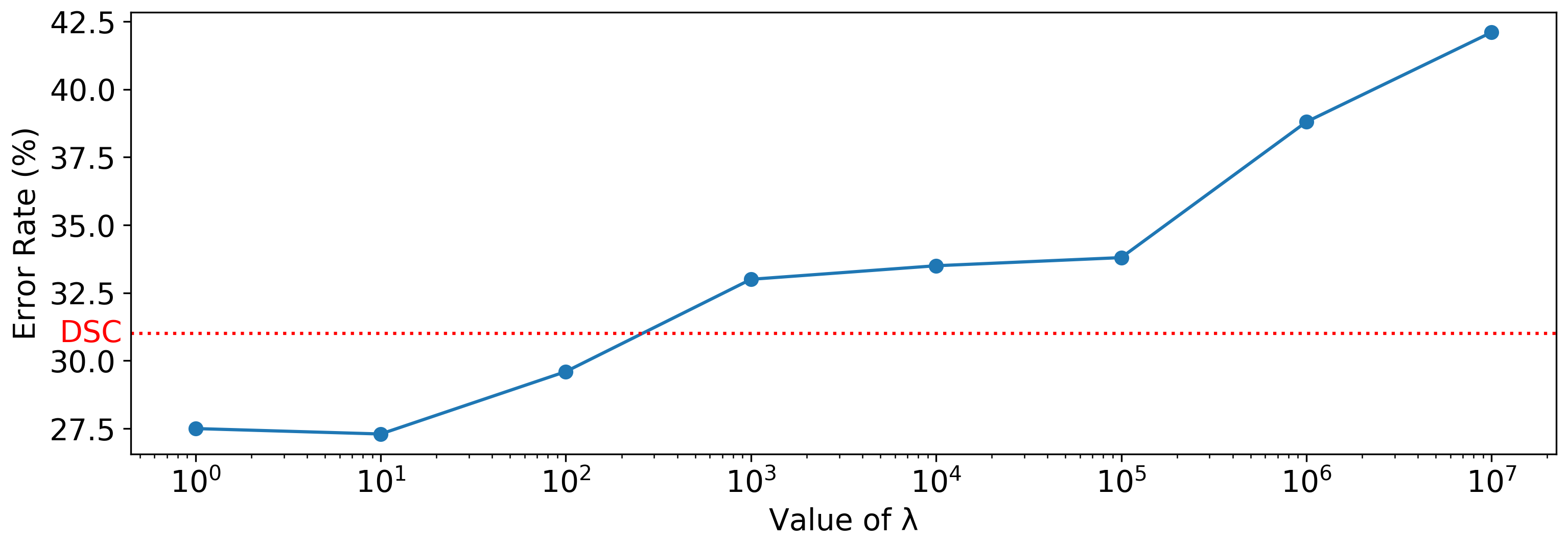}
    \caption{Effects of $\lambda$ on clustering performance of COIL-100 experiments. Performance of DSC baseline is shown as red.
    For all benchmarks, mean of ten trials is reported.}
    \label{fig:effects}
\end{figure}

Figure \ref{fig:effects} shows variation of subspace clustering error in COIL-100 with different selection of $\lambda$.
It can be seen that choosing $\lambda$ from 1 to 100 guaranteed better performance than DSC, and selection from 1 to 10 gave the best performance. Conversely, too large $\lambda$ ($\geq 1e^3$) degenerated performance. On the other hand, with fine-tuning the pre-trained model in small $N$ cases (Section \ref{sec:face}), a relatively large $\lambda$ ($5e^5$) was the appropriate choice for stable convergence. Therefore, the effect of $\lambda$ selection on performance and the following optimal $\lambda$ selection method need to be investigated further in terms of size of dataset, presence or absence of pre-training, and so on.
\section{Conclusions}
In this paper, we firstly propose a variant of the existing DSC method, which does not require retaining parameters of self-expressive layer.
We call our method \textbf{D}eep \textbf{C}losed-\textbf{F}orm \textbf{S}ubspace \textbf{C}lustering (\textbf{DCFSC}) because it is motivated by recently proposed closed form of shallow auto-encoder model.
Our DCFSC has advantages in methodological simplicity and memory efficiency compared to DSC.
Experiments on several benchmarks give two conclusions with regard to DCFSC.
First, the DCFSC model could be trained and converged despite the disadvantage of much less model parameters even in small datasets under the same settings as DSC.
Second, in large dataset, DCFSC could take advantage of memory and eliminate the performance limitations of DSC.
Considering these strengths, we believe that DCFSC can be regarded as a model remedying the shortcomings of the existing DSC.

{\small
\bibliographystyle{ieee_fullname}
\bibliography{egbib}
}

\end{document}